\documentclass{svproc}
\usepackage{epsfig}
\usepackage{svg}
\usepackage{bbm}
\usepackage[algo2e]{algorithm2e}
\usepackage{graphicx}
\usepackage{comment}
\usepackage{booktabs}
\usepackage{algorithm}
\usepackage{graphics}
\usepackage{algpseudocode}
\usepackage{amsmath}
\usepackage{array}
\usepackage{adjustbox}
\usepackage{subcaption}
\usepackage{listings}
\usepackage{float}
\usepackage{enumitem}
\usepackage{xspace}
\usepackage{multirow}
\usepackage{array}
\usepackage{tabularx}
\usepackage{tabu}
\usepackage{bm}
\usepackage{amsfonts}
\usepackage{booktabs, makecell}
\usepackage{xcolor,colortbl}
\usepackage{cleveref}
\usepackage[english]{babel}
\usepackage{pifont} 
\usepackage{soul}
\usepackage{cite}
\usepackage{xfrac}
\usepackage{wrapfig}
\usepackage{mdwlist}
\usepackage{arydshln}
\usepackage[normalem]{ulem}
\usepackage{mathtools}
\newcommand\blfootnote[1]{%
  \begingroup
  \renewcommand\thefootnote{}\footnote{#1}%
  \addtocounter{footnote}{-1}%
  \endgroup
}

\usepackage{url}

\begin{document}
\mainmatter              

\title{Network Design through Graph Neural Networks: Identifying Challenges and Improving Performance}
\titlerunning{Network Design through Graph Neural Networks}

\author{Donald Loveland\inst{1} \and Rajmonda Caceres\inst{2}}
\authorrunning{D. Loveland and R. Caceres} 

\institute{University of Michigan, Ann Arbor MI, USA
\email{dlovelan@umich.edu},
\and MIT Lincoln Laboratory, Lexington MA, USA 
\email{Rajmonda.Caceres@ll.mit.edu}}

\maketitle              

\begin{abstract}
Graph Neural Network (GNN) research has produced strategies to modify a graph's edges using gradients from a trained GNN, with the goal of network design. However, the factors which govern gradient-based editing are understudied, obscuring \textit{why} edges are chosen and \textit{if} edits are grounded in an edge's importance. 
Thus, we begin by analyzing the gradient computation in previous works, elucidating the factors that influence edits and highlighting the potential over-reliance on structural properties.
Specifically, we find that edges can achieve high gradients due to structural biases, rather than importance,  
leading to erroneous edits when the factors are unrelated to the design task.
To improve editing, we propose \textbf{ORE}, an iterative editing method that (a) edits the highest scoring edges and (b) re-embeds the edited graph to refresh gradients, leading to less biased edge choices. 
We empirically study ORE through a set of proposed design tasks, each with an external validation method, demonstrating that ORE improves upon previous methods by up to 50\%.
\end{abstract}
%


\section{Introduction}

Learning over graphs has become paramount in machine learning applications where the data possesses a connective structure, such as social networks \cite{fan2019graph}, chemistry \cite{gilmer2017neural}, and finance \cite{shaksi}. Fortunately, the field of graph mining has provided methods to extract useful information from graphs, albeit often needing heavy domain guidance \cite{liu2019mining}. The advent of graph neural networks (GNNs), a neural network generalized to learn over graph structured data, has helped alleviate some of these requirements by learning representations that synthesize both node and structure information \cite{kipf2016semi, gilmer2017neural, graphsage}. Complimentary to inference, recent work has proposed methods that edit and design network structures using gradients from a trained GNN \cite{lucic2021cfgnn, jin2020graph, li2023interpretable}, enabling the efficient optimization of downstream learning tasks \cite{networkdesign2021} in cyber security \cite{enoch2022cyber, laishram2018}, urban planning \cite{domingo2019roaddesign}, drug discovery \cite{jin2018junction}, and more \cite{ li2022optimal, chan2016, kun2014}. However, as gradient-based editing is applied more broadly, scrutinizing the conditions that allow for successful editing is critical. For instance, discerning the factors which influence gradient computation is still unknown, making it unclear when proposed edits can be trusted. In addition, it is unknown if gradient quality is dependent on graph structure and GNN architecture, causing further concern for practical applications. 

Focusing strictly on gradient-based edit quality, we analyze the common mask learning paradigm~\cite{lucic2021cfgnn, ying2019gnnexplainer, luo2020parameterized, jin2020graph}, where a continuous scoring mask is learned over the edges in a graph. Specifically, we elucidate how structural factors, such as degree, neighborhood label composition, and edge-to-node distance (i.e., how far an edge is from a node) can influence the mask through the gradient. When these factors are not beneficial to the learning task, e.g. edge-to-node distance for a de-noising task when noise is uniformly-distributed across the graph, the learned  mask can lead to erroneous edits.
We additionally highlight how editing methods that rely on thresholding are more susceptible to such structural biases due to smoothing of the ground truth signal at the extreme values of the distribution. To improve editing, we propose
a more fine-tuned sequential editing process, \textbf{ORE}, with two steps: (1) We \textbf{O}rder the edge scores and edit the top-$k$ edges to prioritize high quality edges, and (2) we \textbf{R}e-embed the modified graph after the top-$k$ edges have been \textbf{E}dited. These properties help prevent choosing edges near the expected mask value, and thus more likely to be based on irrelevant structural properties, as well as encourage edits that consider the influence of other removed edges with higher scores. We highlight the practical benefit of ORE by designing a systematic study that probes editing quality across a variety of common GNN tasks, graph structures, and architectures, demonstrating up to a 50\% performance improvement for ORE over previous editing methods.


\section{Related Work}

Early network design solutions choose edits based on fixed heuristics, such as centrality scores\cite{li2022optimal} or triangle closing properties\cite{kun2014}. However, fixed heuristics generally require significant domain guidance and may not generalize to broader classes of networks and tasks. Reinforcement learning (RL) has enabled the ability to learn more flexible heuristics, such as in chemistry \cite{zhou2019optimization} and social networks \cite{morales2021selective}; however, RL can be prohibitively expensive due to data and computation requirements. To fulfill the need for efficient and flexible editing methods,  gradient-based optimization has subsequently been applied to edge editing, facilitated through trained GNNs. Although computing gradients for edges can be infeasible given the discrete nature of the input network, previous methods have adopted a continuous relaxation of the edge set, operating on a soft edge scoring mask that can be binarized to recover the hard edge set~\cite{ying2019gnnexplainer, luo2020parameterized, schlichtkrull2021interpreting, lucic2021cfgnn, jin2020graph}. In its simplest form, the gradient of an edge is approximated as the gradient of the score associated with that edge, with respect to a loss objective \cite{ying2019gnnexplainer}. As this is dependent on the initialization of the scoring mask, GNNExplainer proposes to leverage multiple rounds of gradient descent over the mask to arrive at a final score, rather than use the gradient directly \cite{ying2019gnnexplainer}. CF-GNNExplainer extends GNNExplainer by  generating counterfactual instances and measuring the change in the downstream objective \cite{lucic2021cfgnn}. Both of these methods convert the soft mask to a hard mask through fixed thresholding, which, when incorrectly chosen, can introduce noisy edits. Moreover, as mask learning is usually used to support broader objectives, such as robustness or explainability, studies fail to consider what conditions can inhibit the mask learning sub-component, instead focusing simply on the downstream objective. \textit{Our work provides a direct analysis of mask quality through a systematic study across a wide array of tasks, GNNs, and topologies. We highlight that current mask-based editing methods can become susceptible to bias within the mask scores, prompting the development of ORE as a means of improving gradient-based edge editing}.


\section{Notation}
Let $G = (V, E, \mathbf{X}, \mathbf{Y})$ be a simple graph with nodes $V$, edges $E$, feature matrix $\mathbf{X} \in \mathbb{R}^{|V| \times d}$ with $d$ node features, and label matrix $\mathbf{Y}$. $\mathbf{Y} \in \{0, 1\}^{|V| \times c}$ with $c$ classes for node classification, $\mathbf{Y} \in \mathbb{R}^{|V|}$ for node regression, and $\mathbf{Y} \in \{0, 1\}^{c}$ for graph classification. $\mathbf{A} \in \{0, 1\}^{|V| \times |V|}$ is the adjacency matrix of $G$, where $\mathbf{A}_{i, j} = 1$ denotes an edge between nodes $i$ and $j$ in $G$, otherwise $\mathbf{A}_{i, j} = 0$. While $E$ and $\mathbf{A}$ represent similar information, $E$ is used when discussing edge sets and $\mathbf{A}$ is for matrix computations. Additionally, a \textit{k}-hop neighborhood of a node $i \in V$, $N_{k}(i)$, denotes the nodes and edges that are reachable within $k$-steps of $i$. For simplicity, $k$ is dropped when referring to the 1-hop neighborhood. Additionally, we denote $||\mathbf{B}||_{1}$ as the L$^{1}$-norm of a matrix $\mathbf{B}$, $G - e_{i}$ as the removal of an edge from $G$, and $G - i$ as the removal of a node from $G$. For a $k$-layer GNN, learning is facilitated through message passing over $k$-hop neighborhoods of a graph \cite{gilmer2017neural}. A node $i$'s representations are updated by iteratively aggregating the features of nodes in $i$'s 1-hop neighborhood, denoted AGGR, and embedding the aggregated features with $i$'s features, usually through a non-linear transformation parameterized by a weight matrix $\mathbf{W}$, denoted ENC. The update for node $i$ is expressed as $\mathbf{r}_{i}^{(l)} = \text{ENC}(\mathbf{r}_{i}^{(l-1)}, \text{AGGR}({\mathbf{r}_{u}^{(l-1)}, u \in N(i)}))$ for $l \in \{1, 2, ..., k\}$, where $r_{i}^{(0)} = x_{i}$. The update function is applied $k$ times, resulting in node representations that can be used to compute predictions. For graph-level tasks, a readout function aggregates the final representation of all nodes into a single graph-level representation.

\section{Optimization for Network Editing}

The network design objective is given in Equation \ref{base_opt_prob}, where we want to find a new adjacency matrix, $\mathbf{A^*}$, that improves a function $f$, parameterized by a GNN,

\begin{equation}
	\begin{aligned}
    \min_{\mathbf{A^{*}}} \quad & ||\mathbf{A} - \mathbf{A^{*}}||_{1} \\
    \textrm{s.t.} \quad &  f(\mathbf{X}, \mathbf{A^*}) - f(\mathbf{X}, \mathbf{A}) \geq 0.\\
    \end{aligned}
    \label{base_opt_prob}
\end{equation}
As $\mathbf{A}$ is discrete and $f$ introduces non-linear and non-convex constraints, it is difficult to find an exact solution. Thus, we soften the constraints and focus on increasing $f$ while maintaining the size of $A$, as shown in Equation \ref{relaxed_short},

\begin{equation}
\begin{aligned}
    \min_{\mathbf{A^{*}}} \quad & -f(\mathbf{X}, \mathbf{A^*}) + \lambda ||\mathbf{A} - \mathbf{A^*}||_{1}.
\end{aligned}
\label{relaxed_short}
\end{equation}
where $\lambda$ trades off the objective and the size of the remaining edge set. The negative term incentivizes the optimizer to improve $f$. As the optimization is still over a discrete adjacency matrix, we re-parameterize $\mathbf{A}$, as done in \cite{jang2017categorical, ying2019gnnexplainer}, and introduce a continuous mask $\mathbf{M} \in \mathbb{R}^{n \times n}$. $\mathbf{M}$ is introduced into a GNN's aggregation function as $\textrm{AGGR}({m_{u, v}\cdot \mathbf{r}_{u}^{(i-1)}, u \in N(v)}))$, where $m_{u, v}$ is the mask value on the edge that connects nodes $u$ and $v$. By introducing $\mathbf{M}$ into $\textrm{AGGR}$, it is possible to directly compute partial derivatives over $\mathbf{M}$, enabling gradient-based optimization over the mask values. As the aggregation function is model-agnostic, we can easily inject the mask into any model that follows this paradigm.

\subsection{Graph Properties that Influence Edge Scores}

We aim to study the gradient of the scoring mask $\mathbf{M}$ for a graph $G$. We assume access to a trained, 2-layer GNN with structure $(\mathbf{A}+\mathbf{I})^{2}\mathbf{X}\mathbf{W}$, where $\mathbf{I}$ is the identity matrix. We analyze a node classification setting, where a node $i$'s feature vector is $\mathbf{x}_{i} = \mathbf{y}_{i} + \mathcal{N}(\mu, \Sigma)$, and $\mathbf{y}_{i}$ is the one-hot encoding of class $y_{i}$. After two layers of propagation, the feature vector for node $i$ becomes,

\begin{align}
    \mathbf{r}_{i}^{(2)}  &= \mathbf{x}_{i} + \sum_{j \in N(i)} \mathbf{M}_{i, j} \mathbf{x}_{j} + \sum_{j \in N(i)} \mathbf{M}_{i, j} (\mathbf{x}_{j}+ \sum_{k \in N(j)}\mathbf{M}_{j, k}\mathbf{x}_{k}).
\end{align}

Then, the class prediction for $i$ is $\underset{{z}_{i}}{\operatorname{argmax}}$, where $\mathbf{z}_{i} = \mathbf{r}_{i}^{(2)}W$. As $\mathbf{M}$ is commonly learned through gradient ascent, and only $\mathbf{r}_{i}^{(2)}$ depends on $\textbf{M}$, we focus on the partial derivative of $\mathbf{r}_{i}^{(2)}$ with respect to a mask value $\mathbf{M}_{u, v},$ where $u, v$ are nodes in $G$. 
As the GNN has two layers, the edges must be within two-hops of $i$ to have a non-zero partial derivative. The partial derivative for the one- and two-hop scenarios are the first and second cases of Equation \ref{cases}, respectively,

\begin{align}
\dfrac{\partial \mathbf{r}^{(2)}_{i}}{\partial \mathbf{M}_{u, v}} =
\begin{dcases}
2(\mathbf{y}_{j} + \mathbf{M}_{i, j}\mathbf{y}_{i} + (\mathbf{M}_{i, j} + 1)\mathcal{N}(\mu, \Sigma)) \\
\quad + \sum_{k \in N(j) - i}\mathbf{M}_{j, k}(\mathbf{y}_{k} + \mathcal{N}(\mu, \Sigma)), & u=i, v = j \in N(i) \\
\mathbf{M}_{i, j}(\mathbf{y}_{k} + \mathcal{N}(\mu, \Sigma)), & u = j \in N(i), v = k \in N(j)
\end{dcases}
\label{cases}
\end{align}

To understand the gradient ascent process, we consider when $y_{i} = 0$, without loss of generality, and simplify Equation \ref{cases}. This leads to four scenarios, $y_{j} \in \{0, 1\}$ where $j \in N(i)$ and $y_{k} \in \{0, 1\}$ where $k \in N_{2}(i)$; however, $y_{j}$ only impacts case 1 and $y_{k}$ only impacts case 2, thus we can analyze each in isolation. To elucidate possible biases, we show the difference in gradients by subtracting each possible scenario (for similarly initialized $\textbf{M}_{i, j}$), denoted as $\Delta \partial \mathbf{r}^{(2)}_{i, 0}$, in Equation \ref{diff_cases},

\begin{align}
\Delta \partial \mathbf{r}^{(2)}_{i, 0}
=
\begin{dcases}
(\mathbf{M}_{i, j} + 2)\mathcal{N}(\mu + 1, \Sigma), & y_{j} = 0, y_{k} = 0 \\
\mathbf{M}_{i, j} + (\mathbf{M}_{i, j} + 2)\mathcal{N}(\mu, \Sigma), & y_{j} = 1, y_{k} = 0\\
2(\mathbf{M}_{i, j} + 1) + (\mathbf{M}_{i, j} + 2)\mathcal{N}(\mu, \Sigma), &y_{j} = 0, y_{k} = 1\\
2\mathbf{M}_{i, j} + (\mathbf{M}_{i, j} + 2)\mathcal{N}(\mu, \Sigma), &y_{j} = 1, y_{k} = 1\\
\end{dcases} \nonumber \\
+ \sum_{k \in N(j) - i, y_{k} = y_{j}}M_{j, k}\mathcal{N}(\mu + 1, \Sigma) + \sum_{k \in N(j) - i, y_{k} \ne y_{j}}M_{j, k}\mathcal{N}(\mu, \Sigma).
\label{diff_cases}
\end{align}

First, all cases in Equation \ref{diff_cases} tend to be greater than 0, leading to higher scores for edges closer to $i$. Additionally, if elements of $\mathbf{M} \sim U(-1, 1)$ as in  \cite{ying2019gnnexplainer, lucic2021cfgnn}, the last two summation terms in Equation \ref{diff_cases} scale as $h_{j}(d_{j} - 1)$ and $(1 - h_{j})(d_{j} - 1)$, respectively, where $h_{j}$ and $d_{j}$ represent the homophily and degree properties of the node $j$. Thus, high degree and high homophily can additionally bias edge selection, similar to the heuristic designed by \cite{navigatenetworks} where they use $h_{j}d_{j}$ to optimize network navigation. Each of the above structural factors can either coincide with the true edge importance, or negatively influence edits when such structural properties are uninformative to the network design task.

\subsection{\textbf{ORE}: Improved Edge Editing}
Previous mask learning methods \cite{lucic2021cfgnn, jin2020graph, ying2019gnnexplainer} have focused on fixed thresholding to generate an edge set. As shown above, it is possible that the gradients are biased towards unrelated structural properties, and thus thresholding near the expected mask value can introduce incorrect edits. To improve the mask, we introduce \textbf{ORE}, which operates by sorting the learned mask values, editing only a fixed budget of the highest scoring edges, and then re-embedding the edited graph to obtain an updated mask. Ordering the mask values and only operating on the extreme ends of the mask value distribution allows ORE to choose edges that are likely to be governed by the mask learning procedure, rather than edges with high scores due to structural biases.  Additionally, as seen in Equation \ref{diff_cases}, the gradient for an edge is dependent on downstream edges aggregated during message passing, motivating our re-embedding step to account for interactions between edits. The total editing budget is denoted as $b$, where $b/s$ edges are removed for $s$ steps. If a task requires the solution to contain a single connected component, edges that would disconnect the graph are preserved, their gradients are deactivated, and their mask values are set to one.

\begin{algorithm}

 \KwIn{GNN model $f$, Features $\mathbf{X}$, Adj. Matrix $\mathbf{A}$, Steps $s$, Epochs $e$, Budget $b$, $\lambda$}
 \KwResult{Edited Adjacency Matrix $\mathbf{A}$ }
 Initialize mask matrix $\mathbf{M}$ over edges in $G$ \\
 \For{$1$ to $s$}{
  \For {$1$ to $e$} {
   $P = f(\mathbf{X}, \mathbf{A}, \mathbf{M})$  \tcp*{Forward pass for prediction} 
   $L = -P - \lambda |\mathbf{M}|$ \tcp*{Loss on $P$ (can modify objective)} 
   $\mathbf{M} \leftarrow \mathbf{M} - \alpha \frac{dL}{d\mathbf{M}}$  \tcp*{Update mask} 
  }
  \, $O = \textit{argsort}(\mathbf{M})$ \\
  $I = O[:(b/s)]$ \tcp*{Get top indices to edit}  
  \, $\mathbf{A}[I] = 0$ \tcp*{Remove edges from $G$} 
 }
 Return $\mathbf{A}$ 
 \caption{ORE Algorithm}
 \label{edgeedit}
\end{algorithm}
\raggedbottom


\section{Experimental Setup}

\subsection{Network Editing Process} 

We study four GNN architectures: GCN \cite{kipf2016semi}, GraphSage \cite{graphsage}, GCN-II \cite{chenWHDL2020gcnii}, and Hyperbolic GCN \cite{chami2019hyperbolic}. As attention weights have been shown to be unreliable for edge scoring \cite{ying2019gnnexplainer}, we leave them out of this study. 
After training, each model's weights are frozen and the edge mask variables are optimized to modify the output prediction. We train three independent models on different train-val-test (50-25-25) splits for each task and the validation set is used to choose the best hyperparameters over a grid search. Then, editing is performed over 50 random data points sampled from the test set. For regression tasks, we directly optimize the output of the GNN, and for classification tasks, we optimize the cross entropy loss between the prediction and class label. For ORE, $s = b$ so that one edge is edited per step. Additionally, $b$ is set such that roughly 10\% (or less) of the edges of a graph (or computational neighborhood) are edited. The exact budget is specified for each task. All hyperparameters and implementation details for both the GNN training and mask learning are outlined in an anonymous repo\footnote{\label{anon_repo_impl}https://anonymous.4open.science/r/ORE-93CC/GNN\_details.md}. 

\vspace{.1cm}
\noindent \textbf{Editing Baselines}
We utilize two fixed heuristics for editing: iterative edge removal through random sampling and edge centrality scores~\cite{brandes2001central}. We also study CF-GNNExplainer \cite{lucic2021cfgnn}, though we extend the algorithm to allow for learning objectives outside of counterfactuals and variable thresholds that cause $b$ edits to fairly compare across methods. These changes do not hurt performance and are simple generalizations. Note that while we focus on CF-GNNExplainer, as they are the only previous mask learning work to consider editing, their mask generation is highly similar to other previous non-editing methods, allowing us to indirectly compare to thresholding-based methods in general \cite{ying2019gnnexplainer, luo2020parameterized, schlichtkrull2021interpreting}. 

\subsection{Learning Tasks}

In this section we detail the proposed tasks. For each, the generation process, parameters, and resultant dataset stats are provided in an anonymous repo\footnote{https://anonymous.4open.science/r/ORE-93CC/Dataset\_details\_stats.md}. 

\vspace{.2cm}
\noindent \textbf{Improving Motif Detection:}
We begin with node classification tasks similar to \cite{ying2019gnnexplainer, luo2020parameterized, lucic2021cfgnn} with a goal of differentiating nodes from two different generative models. \textit{Tree-grid} and \textit{tree-cycle} are generated by attaching either a 3x3 grid or a 6 node cycle motif to random nodes in a 8-level balanced binary tree. We train the GNNs using cross entropy, and then train the mask to maximize a node's class prediction. As the generation process is known, we extrinsically verify if an edit was correct by determining if it corresponds to an edge inside or outside of the motifs. The editing budget is set to the size of the motifs, i.e. $b = 6$ for tree-cycle and $b=12$ for tree-grid. Each model is trained to an accuracy of 85\%. 

\vspace{.2cm}
\noindent \textbf{Increasing Shortest Paths (SP):}
The proposed task is to delete edges to increase the SP between two nodes in a graph.
This task has roots in adversarial attacks \cite{miller21} and network interdiction {\cite{smith2013}} with the goal of force specific traffic routes. 
The task is performed on three synthetic graphs: Barabási–Albert (BA), Stochastic Block Model (SBM), and Erdős–Rényi (ER). The parameters are set to enforce each graph has an average SP length of 8. The GNN is trained through MSE of SP lengths, where the SP is estimated by learning embedding for each node and then computing the $L^2$ distance between each node embedding for nodes in the training set. The GNN is then used to increase the SP for pairs of nodes in the test set, which is externally verified through NetworkX. The editing budget $b=30$ given the larger graphs. Each model is trained to an RMSE of 2.

\vspace{.2cm}
\noindent \textbf{Decreasing the Number of Triangles:}
The proposed task is to delete edges to decrease the number of triangles in a graph. Since triangles are often associated with influence, this task can support applications that control the spread of a process in a network, such disease or misinformation \cite{erd2021}. 
We consider the same graphs as in the SP task, BA, SBM, and ER, but instead generate 100000 different graphs each with 100 nodes.
Each generation method produces graphs that, on average, have between 20 and 25 triangles, as computed by NetworkX's triangle counter. The GNNs are trained using MSE and then used to minimize the number of triangles in the graph, which is externally verified through NetworkX. The editing budget $b = 20$. 
Each GNN is trained to an RMSE of 6.

\vspace{.2cm}
\noindent \textbf{Improving Graph-level Predictions:}
MUTAG is a common dataset of molecular graphs used to evaluate graph classification algorithms. The proposed task is to turn mutagenic molecules into non-mutagenic molecules by deleting mutagenic functional groups \cite{luo2020parameterized, ying2019gnnexplainer}. We first train the GNN models to sufficiently predict whether a molecule is mutagenic, then edit the molecules to reduce the probability of mutagenicity. We only edit mutagenic molecules that possess mutagenic functional groups, as in \cite{luo2020parameterized}. The editing budget $b = 5$. Each GNN is trained to an accuracy above 75\%. To focus on edit quality, we do not include chemistry-based feasibility checks, however it is possible to incorporate constraints into ORE either through the mask learning objective, when the constraint is differentiable, or by rejecting edits when the constraint is not differentiable. 


\section{Results}

We present the empirical results for each task, beginning with an in-depth analysis on motif detection. Then, we collectively analyze the shortest path, triangle counting, and mutag tasks, noting trends in editing method and GNN design.

\subsection{Motif Detection}

\begin{figure}[h!]
        \centering
        \includegraphics[width=0.9\textwidth]{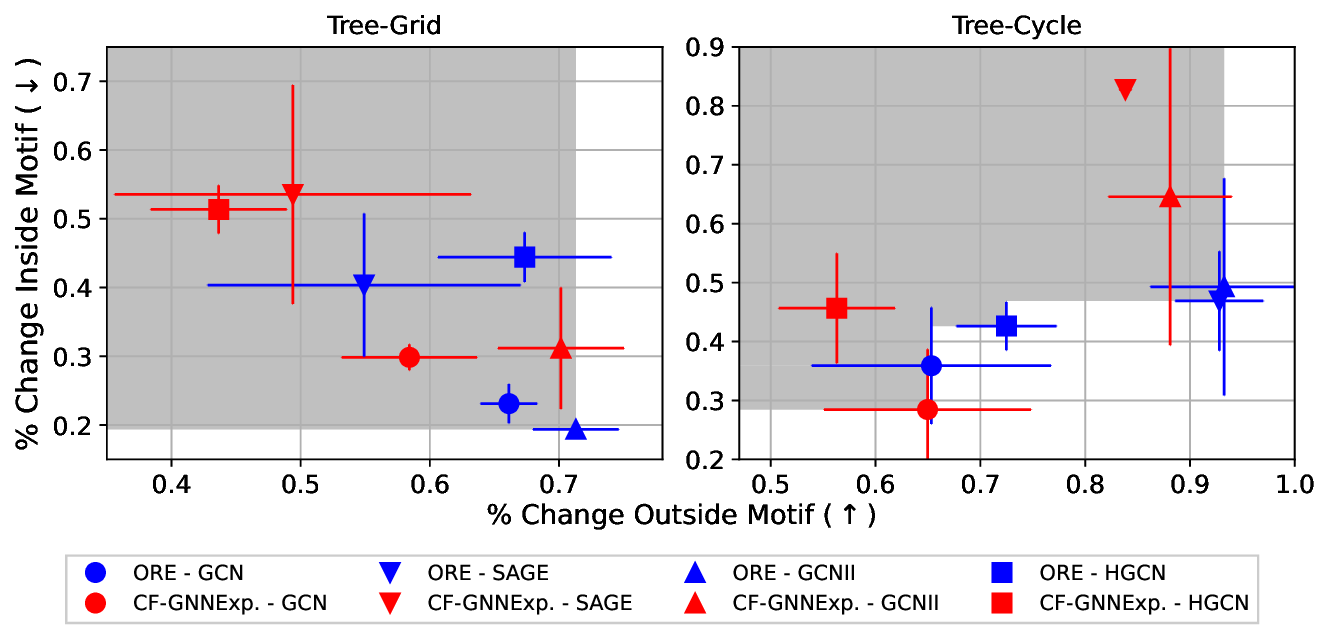}
        \caption{Performance on tree-grid and tree-cycle across GNNs (shapes) and editing methods (colors). The axis show the percent change in edges outside and inside the motifs. Error bars indicate standard deviation in experiments. \textit{Performance improves towards the bottom right}, as the goal is to remove edges outside the motif and retain edges inside the motif, as shown by the gray Pareto front. 
        }
        \label{synth1_global}
\end{figure}

In Figure \ref{synth1_global} we show the percent change metrics for the tree-grid and tree-cycle datasets across the GNN models. Better performance is indicated by a higher percentage of edges removed outside the motif, and a lower percentage of edges removed from inside the motif. We include performance for ORE and CF-GNNExplainer with different GNN backbones. On both datasets, the Pareto front is comprised primarily by variants of ORE, highlighting that ORE is generally better at maintaining in motif edges while removing out of motif edges. 

\vspace{-.4cm}
\vspace{-.2cm}

\begin{figure}[h!]
        \centering
        {\includegraphics[width=.85\textwidth]{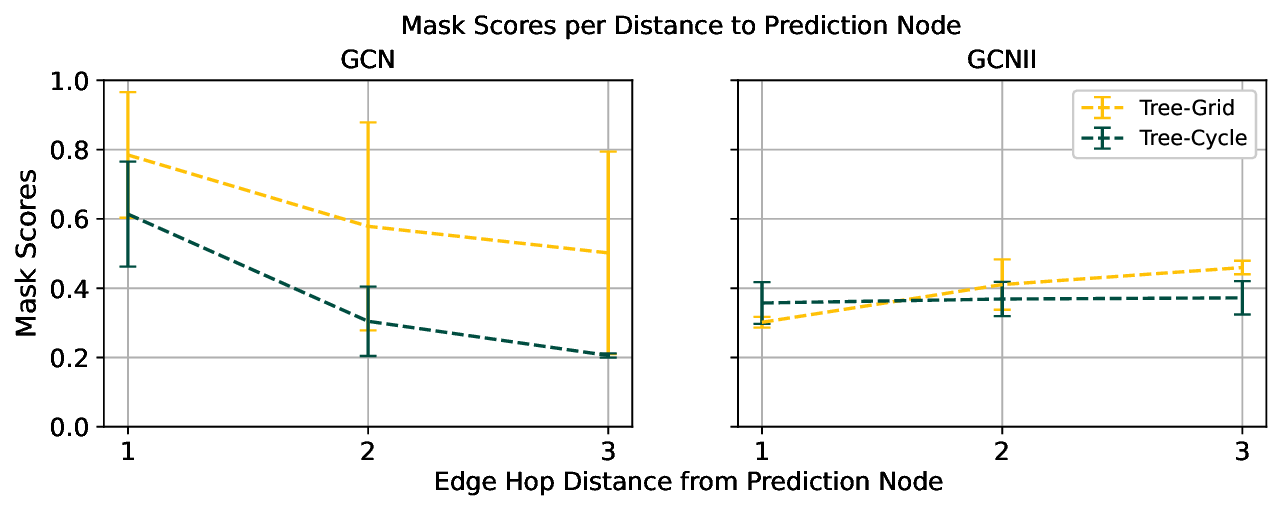}}
        \caption{Mask score distribution stratified by distance to ego-node for \textbf{GCN} and \textbf{GCNII}. Yellow denotes Tree-Grid, green denotes Tree-Cycle. For GCN, the closer an edge is to the ego-node, the higher the scores, leading to bias within the editing. GCNII minimizes bias for this unrelated property, improving editing.}
        \label{mask_cross_models}
\end{figure}

\noindent \textbf{How do editing methods vary across GNNs?}
In Figure \ref{synth1_global}, ORE with GCNII yields the best performance; however, nearly every ORE and GNN combination outperforms the CF-GNNExplainer variant with the same GNN, demonstrating the intrinsic benefit of ORE, as well as the dependence on GNN model. To probe how performance varies across GNNs, we stratify performance by structural factors, as motivated by our analysis in Equation \ref{diff_cases}. In Figure \ref{mask_cross_models}, we focus on the edge-to-node distance, showing that GCN is more susceptible than GCNII to this bias as the correlation between mask score and distance is higher. This result suggests that GCNII is able to suppress the use of factors unrelated to the editing task and better leverage the true importance of the edited edges. We hypothesize that GCNII's ability to retain distinct node representations by combatting oversmoothing can enable more salient gradients, however further theoretical analysis is required to confirm this behavior. 

\vspace{-.3cm}
\vspace{-.2cm}

\begin{figure}[h!]
        \centering
        {\includegraphics[width=0.67\textwidth]{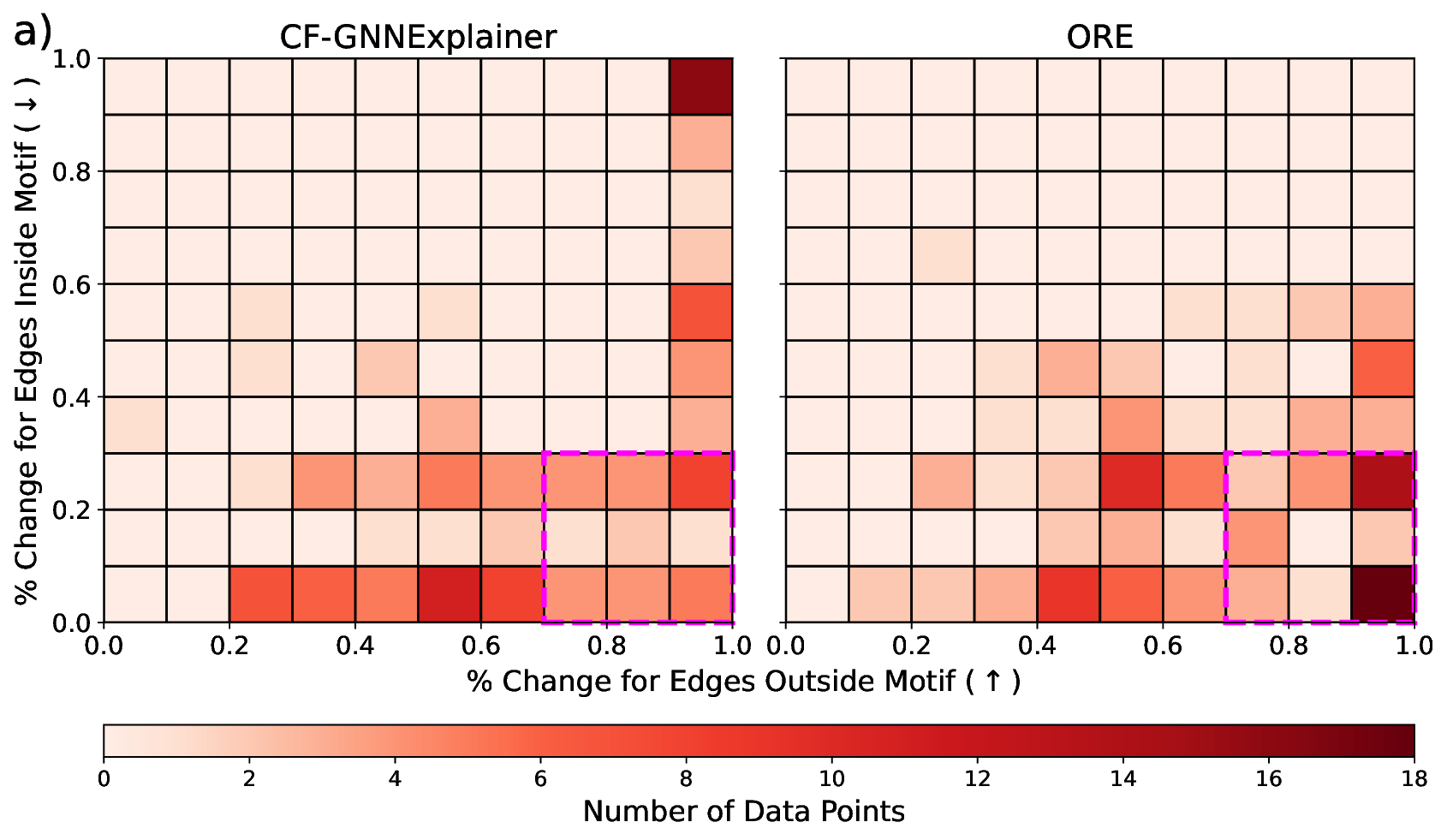}}
        {\includegraphics[width=0.32\textwidth]{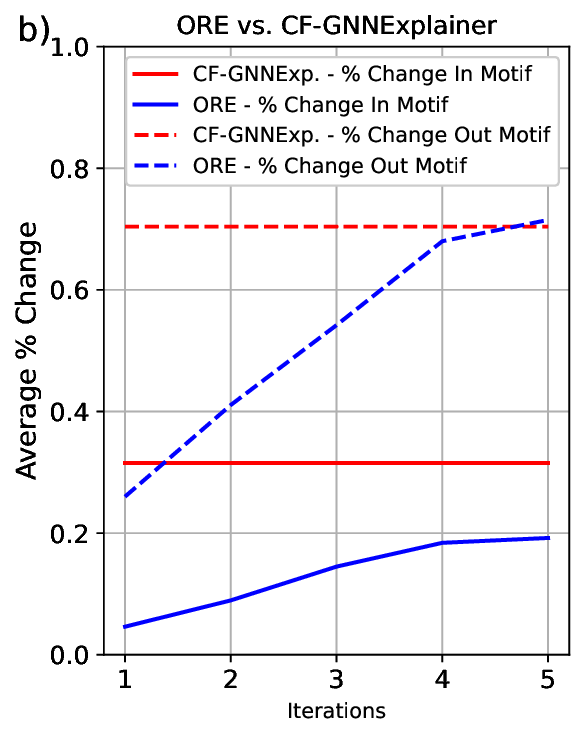}}
        \caption{Analysis on \textbf{GCNII} and \textbf{Tree-Grid}. (a) Histograms where the axes denote the percent change in edges inside and outside of the motif, boxes capture the counts. \textit{ORE outperforms CF-GNNExplainer, as shown by the darker boxes in the bottom right.} (b) Performance across edit iterations. Blue denotes ORE, red denotes CF-GNNExplainer, dashed lines denote out motif change, and solid lines denote in motif change. \textit{ORE rapidly removes edges outside the motifs while maintaining edges inside the motif, improving upon CF-GNNExplainer.} }
        \label{synth1_granular}
        \vspace{-.3cm}
\end{figure}

\noindent \textbf{How does ORE improve performance?}
In Figure \ref{synth1_granular}a, granular performance metrics are presented in a 2D histogram for ORE and CF-GNNExplainer with GCNII, demonstrating the percent change of inside and outside motif edges for tree-grid. Result trends are similar for tree-cycle.
ORE is shown to drop significantly less edges inside the motifs, denoted by the dark red boxes in the bottom right, indicating stronger performance. While both editing methods perform well at removing edges outside the motifs, CF-GNNExplainer tends to additionally remove inside edges, indicating a poorer trade-off between outside and inside motif edges. We further analyze how this arises in Figure \ref{synth1_granular}b, where the percent change metrics are presented across edit iterations (CF-GNNExplainer is not iterative and thus constant). For ORE, we see that the rates of change for inside and outside edges are significantly different -- ORE more rapidly removes outside edges while maintaining inside edges, improving the final edit solution. In addition, ORE achieves similar outside edge removal to CF-GNNExplainer, while achieving a 10\% increase in inside edges, supporting our hypothesis that knowledge of earlier edits allows ORE to adjust mask scores, improving editing.

\subsection{Shortest Path, Triangle Counting, and Graph Classification}

In Table \ref{edit_perf}, we outline the performance metrics for the SP, triangle counting, and mutag tasks. For each task, we measure the average percent change in their associated metric. In the SP experiments, all GNNs improve over the baselines, demonstrating the learned masked values extracted from the GNNs can outperform crafted heuristics, such as centrality, which leverages shortest path information in its computation. Given that ORE with GCN performs well on this task, it is possible that the structural biases identified previously, such as reliance on degree, could coincide with the SP task and improve mask scores. In the triangle counting task, edge centrality is a strong baseline for BA graphs, likely due to centrality directly editing the hub nodes that close a large number of triangles. Across the ER and SBM graphs, which do not possess a hub structure, we find that ORE with a GCNII backbone performs significantly better than both the baselines and other GNN models. Mutag reinforces these findings where GCNII removes nearly all of the mutagenic bonds for the mutagenic molecules. Notably, the Hyperbolic GCN performs poorly across experiments, possible explained by most tasks possessing Euclidean geometry, e.g. 82\% of the molecules in the mutagenic dataset are roughly Euclidean as computed by the Gromov hyperbolicity metric \cite{gromov}. Comparing editing methods, ORE with GCN and GCNII significantly outperforms CF-GNNExplainer with GCN across all three downstream tasks, highlighting the value of refined and iteratively optimized edge masks.

\begin{table}[t!]
\centering
\caption{Results for SP, triangle counting, and mutag tasks. CF-GNNExplainer leverages a GCN, often one of the better performers in motif analysis. All metrics are average percent change, where higher is better. Error is the standard deviation across each model. The highlighted boxes indicate best performaners.}
{\fontsize{6.8}{9.2}\selectfont
\begin{tabular}{@{}llll|lll|l@{}}
\toprule
               & \multicolumn{3}{l|}{Shortest Path} & \multicolumn{3}{l|}{Triangle Counting} & \multicolumn{1}{l}{Mutag}   \\ \midrule
               & BA     & ER     & SBM     & BA         & ER         & SBM       & --- \\ \midrule
Random&$0.0 \pm 0.0$&$0.0 \pm 0.0$&$0.0 \pm 0.0$&$0.36 \pm 0.02$&$0.29 \pm 0.01$&$0.38 \pm 0.0$ & $0.48 \pm 0.04$\\
Centrality&$0.04 \pm 0.01$&$0.08 \pm 0.02$&$0.13 \pm 0.0$&$0.59 \pm 0.02$&$0.07 \pm 0.0$&$0.09 \pm 0.01$ & $0.70 \pm 0.07$ \\\midrule
CF-GNNEx & $0.26 \pm 0.05$ & $ 0.31 \pm 0.04$ & $0.23 \pm 0.03$ & $0.42 \pm 0.04$ & $0.24 \pm 0.09$ & $0.32 \pm 0.06$ & $0.72 \pm 0.04$ \\ \midrule
ORE-GCN&\cellcolor{blue!25}$0.52 \pm 0.06$&\cellcolor{blue!25}$0.89 \pm 0.05$&\cellcolor{blue!25}$0.33 \pm 0.06$&$0.57 \pm 0.05$&$0.33 \pm 0.04$&$0.36 \pm 0.06$ & $0.66 \pm 0.08$ \\
ORE-SAGE&$0.36 \pm 0.11$&$0.74 \pm 0.07$&$0.17 \pm 0.07$&$0.37 \pm 0.05$&$0.29 \pm 0.04$&$0.37 \pm 0.06$ & $0.42 \pm 0.17$\\
ORE-GCNII&$0.24 \pm 0.04$&$0.47 \pm 0.11$&$0.29 \pm 0.06$&\cellcolor{blue!25}$0.64 \pm 0.05$&\cellcolor{blue!25}$0.41 \pm 0.04$&\cellcolor{blue!25}$0.52 \pm 0.05$ & \cellcolor{blue!25}$0.89 \pm 0.04$\\
ORE-HGCN&$0.38 \pm 0.06$&$0.73 \pm 0.06$&$0.18 \pm 0.05$&$0.36 \pm 0.08$&$0.40 \pm 0.04$&$0.45 \pm 0.07$ & $0.47 \pm 0.16$\\\bottomrule
\end{tabular}
}
\label{edit_perf}
\end{table}


\section{Conclusion}

In this work, we focused on studying network design though gradient-based edge editing. We began by identifying structural factors that influence the common mask-based learning paradigm, and empirically demonstrated how these factors can impact performance across complex models and tasks. To improve editing, we introduced a sequential editing framework, ORE, that allowed for (a) the identification of higher quality edges near the extremes of the mask distribution and (b) mask scores to reflect updates from higher scoring edges. As network design evaluation has limited datasets, we proposed a set of editing tasks with external validation mechanisms, and studied both ORE and a strong editing baseline, CF-GNNExplainer, with different GNN backbones. We found that ORE outperformed CF-GNNExplainer across all experiments, while additionally demonstrated the impact of GNN architecture on the success of editing.

\vspace{-.2cm}

%
%
\bibliographystyle{plain} 
\scriptsize{
\bibliography{main} 
}
\vspace{-0.3cm}
\blfootnote{\scriptsize Distribution Statement A. Approved for public release. Distribution is unlimited. This material is based upon work supported by the Under Secretary of Defense for Research and Engineering under Air Force Contract No. FA8702-15-D-0001. Any opinions, findings, conclusions or recommendations expressed in this material are those of the author(s) and do not necessarily reflect the views of the Under Secretary of Defense for Research and Engineering. © 2023 Massachusetts Institute of Technology. Delivered to the U.S. Government with Unlimited Rights, as defined in DFARS Part 252.227-7013 or 7014 (Feb 2014). Notwithstanding any copyright notice, U.S. Government rights in this work are defined by DFARS 252.227-7013 or DFARS 252.227-7014 as detailed above. Use of this work other than as specifically authorized by the U.S. Government may violate any copyrights that exist in this work.}

\end{document}